\documentclass[10pt,twocolumn,letterpaper]{article}

\usepackage{cvpr}
\usepackage{times}
\usepackage{epsfig}
\usepackage{graphicx}
\usepackage{amsmath}
\usepackage{amssymb}
\usepackage{bm}
\usepackage{multirow}
\usepackage{makecell}
\usepackage{wrapfig}
\usepackage{enumitem}
\usepackage{euscript}
\usepackage[breaklinks=true,bookmarks=false]{hyperref}
\setlist[itemize]{leftmargin=*}

\cvprfinalcopy 

\def\cvprPaperID{****} 

\newcommand \V {\mathcal{V}}
\newcommand \E {\mathcal{E}}

\newcommand \tc {\tilde{c}}
\newcommand \cs {c_{sel}}

\begin{document}


\title{Optimal Strategies for Matching and Retrieval Problems by Comparing Covariates}

\author{Yandong Wen, Mahmoud Al Ismail,   Bhiksha Raj, Rita Singh \\
Carnegie Mellon University, Pittsburgh, PA, USA\\
{\tt\small yandongw@andrew.cmu.edu, mahmoudi@andrew.cmu.edu}
}

\maketitle

\begin{abstract}
In many retrieval problems, where we must retrieve one or more entries from a gallery in response to a probe, it is common practice to learn to do by directly
comparing the probe and gallery entries to one another. In many situations
the gallery and probe have common covariates -- external variables that are 
common to both. In principle it is possible to perform the retrieval based
merely on these covariates. The process, however, becomes gated by our ability
to recognize the covariates for the probe and gallery entries correctly.

In this paper we analyze optimal strategies for retrieval based only on
matching covariates, when the recognition of the covariates is itself inaccurate.
We investigate multiple problems:  recovering one item from a gallery of $N$ entries, matching pairs of instances, and retrieval from large collections. We verify our analytical formulae through experiments to verify their correctness in practical settings.
\end{abstract}

\section{Introduction}
{\bf{\emph This document is currently incomplete, and has been uploaded primarily as a supporting document for \cite{wen2018disjoint}.  The completed version will be uploaded shortly}}

Consider the following problem: we are given a ``gallery'' of $N$ items and a single ``probe'' entry, which we expect ``matches'' some of the g entries, in some sense.  Our task is to retrieve the gallery entries that match the probe.

A typical problem, for instance, is when we are given a gallery of $N$ biometric identifiers, such as faces, and a probe instance, which is also a biometric instance (e.g. another face image, or even any other modality such as fingerprint or voice). We must retrieve the appropriate gallery entries that are from the same person as the probe entry.  Alternately, we may be given a gallery of documents by a number of authors, and a probe document of an unknown author.  We must find the gallery entries that match the probe. Many other such problems can be found.

In these problems, the general solution is to find statistical dependencies between entries that relate the two types of data (from the probe and the gallery) and recover the matching entries based on these. Typically, the solution comprises considering some variant of $Prob(match | probe,\ gallery\ entry)$, or $Prob(probe,\ gallery\ entry | match)$ for each of the gallery entries, and determining the match based on this value \cite{webb2003statistical}. This probability itself may utilize any kind of underlying statistical model. These joint models must often be learned from joint presentation of the types of data present in the probe and the gallery.

Often, however, we can find common {\em covariates} to the probe and gallery data, which can be independently determined. For instance, in biometric identification, the {\em gender}, {\em ethnicity}, {\em nationality}, and even characteristics such as body size, affect both the probe and gallery entries. In the document case, the gender and nationality of the author, the writing style, etc., affect the probe and gallery entries. 

We expect the covariate values for probe and its matching gallery entries to be identical.

The key feature here is that these covariates, being known entities, may be independently determined from both probe and gallery entries. For instance, in the biometric problem the gender or ethnicity of a subject may be independently determined for both of them. In the document problem, the gender, nationality, and writing style of the author can be independently determined for the probe and gallery entries. To learn to identify these covariate characteristics, the {\em joint} distribution of probe and gallery data need not be considered at all.

The question we address here how {\em accurate} can retrieval be when it is based only on matching the covariate information of probe and gallery entries, i.e., if the only information used for the retrieval was the estimated covariate values of the probe and gallery data. E.g., in recovering the correct face from a gallery, how accurate would we be if all we did was to match the estimated gender of the probe and gallery entries.

We analyze this problem in a number of different settings.
\begin{itemize}
\item  {\em Retrieval of unique match from a gallery of $N$}. Here we assume a gallery of $N$ entries, where exactly {\em one} of the gallery entries matches the probe.
\item {\em Verification}. The gallery comprises only one entry. I.e., given two data instances, one nominally the probe, and the other the gallery, we must determine if they both match or not.

\end{itemize}
In all of these settings we will derive an optimal policy for identifying gallery matches to the probe, and the error to be expected in following this policy.

We will make some simplifying assumptions. We assume that the gallery entries are all independently drawn and do not inform about one another. We assume only a single probe entry.  Also, although we assume that only a single covariate is considered at any time; this is not a real restriction -- groups of covariates fall into the same analysis by simply considering the group as a single extended covariate. 

We will assume that the ``imposter'' entries in the gallery, {\em i.e.} gallery entries that do not match the probe, are drawn independently of the probe.  We will also assume that the errors in determining the covariate values of gallery entries  are independent of the errors made on probe.

\section {Retrieval of a Unique Match from a Gallery Of $N$} \label{sec:1ofN}
We first consider the problem of retrieval from a gallery of $N$ entries, where it is known that exactly one of the entries is a match to the probe.

Consider a covariate $C$ that can take values in the set $\V_C$.  We will assume for this document that $\V_C$ is discrete, although this is not necessary.

\begin{figure}[t]
  \centering
  \setlength{\abovecaptionskip}{-1.5pt}
  \setlength{\belowcaptionskip}{-14pt}
  \includegraphics[width=3in]{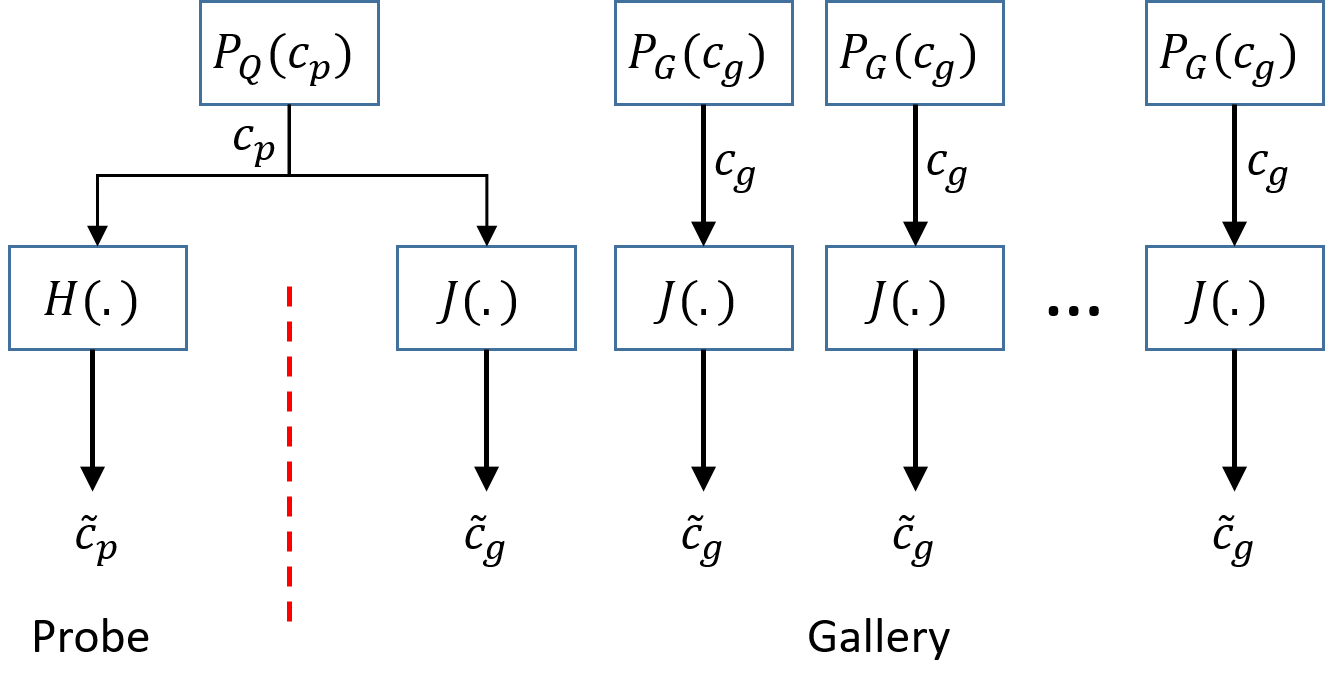}
  \caption{\footnotesize Noisy channel model for the probe and gallery}\label{fig:channelmodel}
\end{figure}

Figure \ref{fig:channelmodel} displays our model. We model the automatic classification of the covariate values for probe instances as a noisy channel $H()$. We model the automatic classification of the covariate values for gallery instances as a noisy channel $J()$. 

The overall model has the following statistical components:
\begin{itemize}
\item A probability distribution $P_Q(c_p)$ from which probe entries with covariates $c_p \in \V_C$ are selected.
\item A probability distribution $P_H(\tc_p|c_p),\ \tc_p\in\V_C$ which specifies the probability that the noisy channel $H()$ will ouput the value $\tc_p$ in response to input $c_p$.
\item A probability distribution $P_G(c_g)$ from which ``imposter'' gallery entries $c_g \in \V_C$ are drawn. 
\item A probability distribution $P_J(\tc_g|c_g),\ \tc_g\in\V_C$ which specifies the probability that the noisy channel $J()$ will ouput the value $\tc_g$ in response to input $c_g$.
\end{itemize}
We assume all of these distributions are known.

The generative model for the process is as follows:
\begin{itemize}
\item  The probe entry $c_p \in \V_C$ is drawn according to probability $P_Q(c_p)$. 
\item $c_p$ is passed through the channel $H()$, which outputs a noisy covariate $\tc_p$ in response.
\item The probe entry $c_p$ is passed through the channel $J()$, which outputs the noisy covariate $\tc_g \in \V_C$ in response. $\tc_g$ is added to the gallery as the matching entry to the probe. 
\item To fill the rest of the gallery of size $N$,  $N-1$ additional entries with covariates $c_g \in \V_C$ are drawn independently according to $P_G(c_g)$. Each of these is passed through the channel $J()$ to obtain the noisy covariate $\tc_g$, which is included in the gallery.
\end{itemize}

Note that using the known distributions, we can also compute the following terms:
\begin{itemize}
\item The overall probability of observing a noisy probe value $\tc_p$:
\[
P_{QH}(\tc_p) = \sum_{c_p} P_Q(c_p) P_H(\tc_p | c_p)
\]
\item The {a posteriori} probability of the true probe covariate $c_p$, given the noisy value $\tc_p$
\[
P_{QH}(c_p | \tc_p) = \frac{P_Q(c_p) P_H(\tc_p | c_p)}{P_{QH}(\tc_p)}
\]
\item The overall probability that any particular gallery item (other than the entry matching the probe) will take a specific value $\tc_g$. From the above formulation, we have
\[
P_{GJ}(\tc_g) = \sum_{c_g} P_G(c_g) P_J(\tc_g | c_g)
\]
and
\[
P_{GJ}(c_g | \tc_g) = \frac{P_G(c_g) P_J(\tc_g | c_g)}{P_{GJ}(\tc_g)}
\]
\end{itemize}

\subsection{Defining a Policy for the Matching}
We will consider a stochastic policy where, given an output $\tc_p$ from $H()$, we select a covariate value $\cs \in \V_C$ according to a probability distribution $r(\cs|\tc_p)$, and subsequently select one of the gallery entries for which $\tc_g = \cs$.  Note that this is a generalization of the more conventional {\em deterministic} policy (which would return a unique $\cs$ in response to each $\tc_p$. As we shall see, however, the optimal strategy is indeed deterministic).

\subsection{Probability of Correctness as a Function of Policy} \label{ssec:correctness}

If there are $K$ gallery entries for which $\tc_g = \cs$, then the probability of correctly matching the probe, given that the original probe entry was $c_p$, is given by
\begin{align}
P(correct |c_p, \cs, K) = P_J(\cs|c_p)\frac{1}{K}.
\end{align}
This factors in both, the probability that the output of the noisy channel $J()$ in response to $c_p$ is $\cs$, and that we have chosen the correct instance from the $K$ gallery items for which $\tc_g = \cs$.

The probability that exactly $K$ of the gallery items will have value $\cs$, given that the matching entry is also $\cs$ is given by
\begin{equation}
P(K) = B(N-1,K-1,P_{GJ}(\cs)),
\label{eq:pk}
\end{equation}
where $B(N,K,p)$ is the binomial probability or choosing $K$ of $N$ entries, with probability parameter $p$:
\[
B(N,K,p) = {\binom{N}{K}}p^{K} (1- p)^{N-K}.
\]
Equation \ref{eq:pk} considers the fact that if we are given that one of the $K$ is the matching entry, we must only account for the ways in which $K-1$ of the remaining $N-1$ gallery entries can also be $\cs$.

The overall probability of correctness of the response when we choose $\cs$ is 
\begin{align}
P(correct |c_p, \cs) = P_J(\cs|c_p)\sum_{K=0}^{N-1} P(K)\frac{1}{K},\nonumber \\
= P_J(\cs|c_p)\sum_{K=1}^{N}  \frac{B(N-1,K-1,P_{GJ}(\cs))}{K}, \nonumber \\
= \frac{P_J(\cs|c_p)}{NP_{GJ}(\cs)}\sum_{K=1}^{N}  B(N,K,P_{GJ}(\cs)) \nonumber \\
 = \frac{P_J(\cs|c_p)(1 - P_{GJ}(\cs))^N}{NP_{GJ}(\cs)}.
\end{align}

Using the law of iterated expectations we can now write the overall probability of correctness of the response, given a noisy probe $\tc_p$ as
\begin{align}
P(correct | \tc_p) &= \E_{\cs|\tc_p} \E_{c_p|\tc_p} P(correct |c_p, \cs) \nonumber \\
&= \sum_{\cs} r(\cs | \tc_p) \sum_{c_p} P_{QH}(c_p|\tc_p) \nonumber \\
& \frac{P_J(\cs|c_p)(1 - P_{GJ}(\cs))^N}{NP_{GJ}(\cs)}.
\label{eq:pcorr}
\end{align}

\subsection{Optimal Policy}

Our objective is to find the policy that maximizes the probability of correctness for any probe $\tc_p$:
\[
\arg\max_{\{r(\cs|\tc_p)\}} P(correct | \tc_p)
\]

Define $\hat{c}(\tc_p)$ as
\[
\hat{c}(\tc_p) = \arg\max_{\cs} \left( \sum_{c_p} P_{QH}(c_p|\tc_p)  \frac{P_J(\cs|c_p)(1 - \tilde{P}_G(\cs))^N}{N\tilde{P}_G(\cs)} \right)
\]

From inspection of Equation \ref{eq:pcorr} we obtain the following optimal policy.
\begin{equation}
r(\cs | \tc_p) = 
\begin{cases}
1,\ \ \cs = \hat{\cs}(\tc_p) \\
0\ \ {\rm else}
\end{cases}.
\label{eq:optrule}
\end{equation}

\subsection{Optimal Error}
Given any noisy probe $\tc_p$, the probability error under the optimal policy is given by
\begin{align}
&P_{opt}(error|\tc_p) = 1 - \nonumber \\
&\max_{\cs} \left( \sum_{c_p} P_{QH}(c_p|\tc_p)  \frac{P_J(\cs|c_p)(1 - \tilde{P}_G(\cs))^N}{N\tilde{P}_G(\cs)} \right)
\label{eq:opterrorgivenprobe}
\end{align}

The overall probability of error is given by
\begin{align}
&P_{opt}(error) = 1 - \sum P_{QH}(\tc_p) P_{opt}(error|\tc_p) \nonumber \\
&= 1 - \nonumber\\
&\max_{\cs} \left(\sum_{c_p, \tc_p} P_Q(c_p)  P_H(\tc_p | c_p) \frac{P_J(\cs|c_p)(1 - \tilde{P}_G(\cs))^N}{N\tilde{P}_G(\cs)}\right)&
\label{eq:opterror}
\end{align}

\section{The Verification Problem}
\begin{figure}[t]
  \centering
  \setlength{\abovecaptionskip}{-1.5pt}
  \setlength{\belowcaptionskip}{-14pt}
	\includegraphics[width=3in]{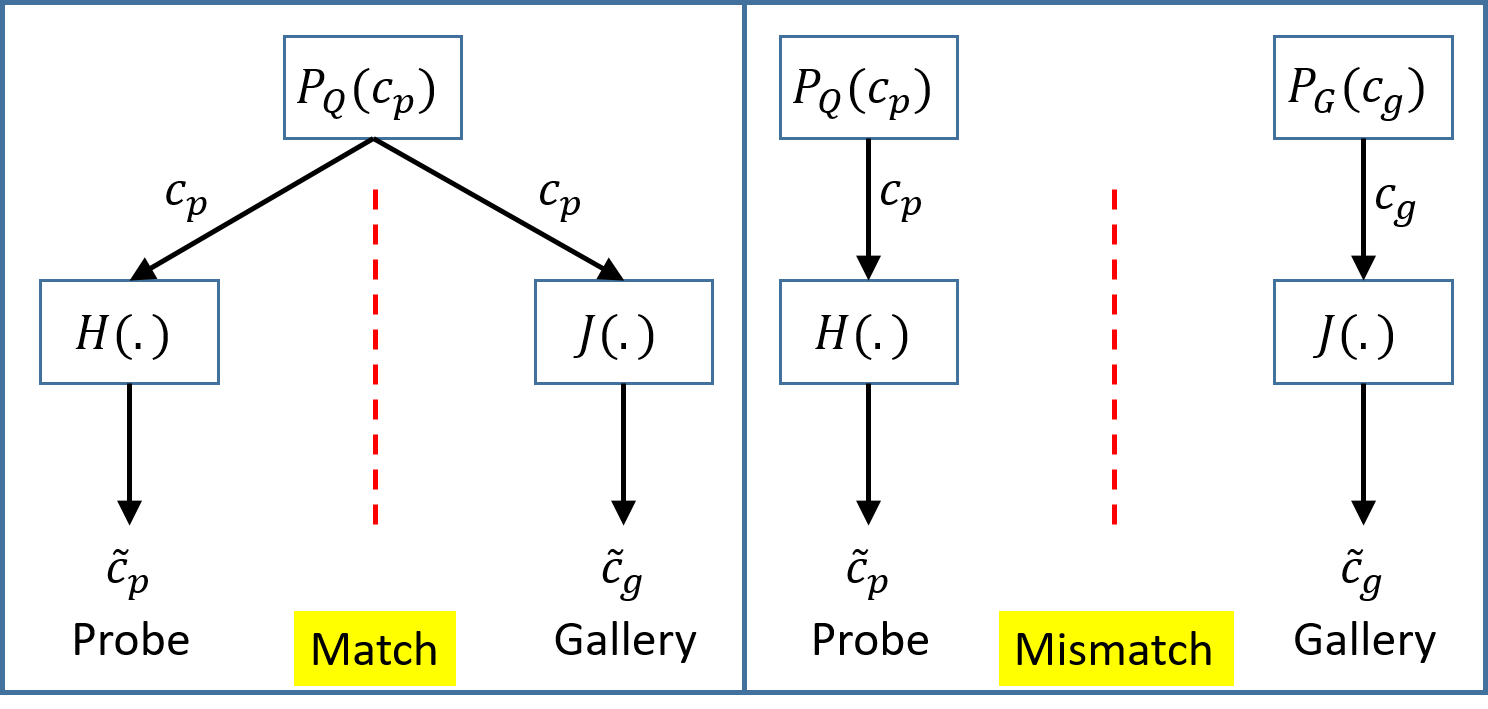}
  \caption{\footnotesize Noisy channel model for the verification problem}\label{fig:verificationchannelmodel}
\end{figure}

Figure \ref{fig:verificationchannelmodel} shows our model for the verification problem. We have two conditions: ``match'' and ``mismatch''.  Under match, a single covariate $c_p$ is drawn from $P_Q(c_p)$ and passed through the two noisy channels $H()$ and $J()$ to produce the probe entry $\tc_p$ and the gallery entry $\tc_g$.  Under mismatch, $c_p$ and $c_g$ are drawn independently from $P_Q(c_p)$ and $P_J(c_g)$ respectively and passed through $H()$ and $J()$ to produce $\tc_p$ and $\tc_g$.

From observing $\tc_p$ and $\tc_g$ we must determine which of the two conditions,  $match$ or $mismatch$, produced them.

\subsection{Defining The Error}
To analyze the problem we must first define the error of matching apporpriately.

When we wrongly identify a case of match as a mismatch ({\em i.e.} we ``reject'' a match), we have an instance of a {\em false rejection}. When a mismatch is misidentified as a match ({\em i.e.} we ``accept'' a mismatch), we have a {\em false acceptance}.  

Let $F_R$ represent the ``false rejection rate'', {\em i.e.} the probability that a match will be wrongly rejected. Let $F_A$ represent the ``false acceptance rate'', i.e. the probability that a negative match will be wrongly accepted. Any classifier can generally be optimized to trade off $F_R$ against $F_A$. The ``Equal Error Rate'' (EER) is achieved when $F_R = F_A$, {\em i.e.} $EER = F_A$ (or $F_R$) when $F_R = F_A$.

We will choose as our objective the minimization of the EER. Note that if an operating point other than EER is chosen to quantify performance ({\em e.g.} $F_A = \beta F_R$ for $\beta \neq 1$, or for some fixed $F_A$ or $F_R$), the analysis below can generally be modified to accommodate it, provided a feasible solution exists.

\subsection{Defining a Policy}

We will use the following stochastic policy: for any pair of noisy probe and gallery values, $\tc_p$ and $\tc_g$, we will accept the pair as a match with probability $r(\tc_p, \tc_g)$.  We must find the $r()$ that minimizes the EER.

\subsection{Error as a Function of Policy}
We first define the probabilities of observing any given $(\tc_p, \tc_g)$ pair under conditions of match and mismatch. From the model of Figure \ref{fig:verificationchannelmodel} we get the following probability under match:
\begin{align}
P(\tc_p, \tc_g |match) &= \sum_{c_p} P(\tc_p, \tc_g, c_p |match), \nonumber \\
&= \sum_{c_p} P_H(\tc_p |c_p)P_J(\tc_g | c_p) P_Q(c_p).
\end{align}
Above we're utilizing the fact that $\tc_p$ and $\tc_g$ are conditionally independent of $match$, given $c_p$. 

Similarly, from Figure \ref{fig:verificationchannelmodel}, the probability of any $(\tc_p,\tc_g)$ under mismatch is given by
\begin{align}
P(\tc_p, \tc_g |mismatch) &= \sum_{c_p,c_g} P(\tc_p, \tc_g, c_p, c_g |mismatch) \nonumber \\
&= \sum_{c_p, c_g} P_H(\tc_p |c_p) P_J(\tc_g|c_g) P_Q(c_p) P_J(c_g)
\end{align}

The probability of a false acceptance is given by
\begin{align}
F_A &= P(accept | mismatch) \nonumber \\
&= \sum_{\tc_p, \tc_q}r(\tc_p, \tc_g) P(\tc_p, \tc_g |mismatch) 
\end{align}

The probability of a false rejection is given by
\begin{align}
F_R &= P(reject | match) \nonumber \\
&= \sum_{\tc_p, \tc_q}(1 - r(\tc_p, \tc_g)) P(\tc_p, \tc_g |match) 
\end{align}

We obtain EER when $F_A = F_R$, {\em i.e.}
\begin{align}
&\sum_{\tc_p, \tc_q}r(\tc_p, \tc_g) P(\tc_p, \tc_g |mismatch) = \sum_{\tc_p, \tc_q}(1 - r(\tc_p, \tc_g)) P(\tc_p, \tc_g |match) \nonumber \\
&\Longrightarrow \sum_{\tc_p, \tc_g} (P(\tc_p, \tc_g | mismatch)+P(\tc_p, \tc_g | match))r(\tc_p, \tc_g) = 1
\end{align}

Thus, optimizing the policy requires solving the following
\[
\begin{split}
&\arg\min_{\{r(\tc_p, \tc_g)\}} \sum_{\tc_p, \tc_q}r(\tc_p, \tc_g) P(\tc_p, \tc_g |mismatch) \nonumber\\
&{\mathrm s.t.} 1\geq r(\tc_p, \tc_g)\geq 0,\nonumber \\
&\sum_{\tc_p, \tc_g} (P(\tc_p, \tc_g | mismatch)+P(\tc_p, \tc_g | match))r(\tc_p, \tc_g) = 1. 
\end{split}
\]

{\small
\bibliographystyle{ieee}
\bibliography{egbib}
}

\end{document}